\documentclass[letterpaper,10pt,conference]{ieeeconf}
\usepackage{float}
\usepackage{cite}
\IEEEoverridecommandlockouts
\usepackage[T1]{fontenc}
\usepackage{booktabs}
\usepackage{scalerel}
\usepackage{medmath}

\usepackage{algorithm}
\usepackage{algpseudocode}
\usepackage{amsmath,amssymb,amsthm,mathtools}
\let\labelindent\relax
\usepackage{microtype,enumitem,booktabs,array,float,balance}
\newtheorem{assumption}{Assumption}
\newtheorem{problem}{Problem}
\newtheorem{theorem}{Theorem}
\newtheorem{lemma}{Lemma}
\newtheorem{corollary}{Corollary}

\theoremstyle{remark}

\floatstyle{ruled}
\newcommand{\mc}{\mathcal}
\newfloat{algorithm}{t}{loa}
\floatname{algorithm}{Algorithm}
\usepackage[most]{tcolorbox}
\usepackage{mdframed}
\usepackage{xcolor}
\newcommand{\smallsquare}{\scalebox{0.7}{$\blacksquare$}}

\definecolor{mygreen}{rgb}{0.0, 0.3, 0.0}
\definecolor{winered}{rgb}{0.8,0,0}
\definecolor{myblue}{rgb}{0,0,0.8}
\usepackage[colorlinks]{hyperref}
\hypersetup{
  citecolor=myblue,
  linkcolor=winered
}

\newcommand{\R}{\mathbb R}
\newcommand{\E}{\mathbb E}

\newcommand{\norm}[1]{\lVert #1\rVert}

\newcommand{\sat}{\textcolor{winered}{\texttt{sat}}}
\newcommand{\gbar}{\bar g}

\newcommand{\aqtd}{\textnormal{\texttt{BAQ-TD}}}
\newcommand{\td}{\textnormal{TD~}}
\setlist[enumerate]{leftmargin=*,itemsep=2pt,topsep=3pt}
\setlist[itemize]{leftmargin=*,itemsep=2pt,topsep=3pt}
\allowdisplaybreaks[1]

\title{\LARGE\bfseries Adaptive Quantization Preserves Fast Rates for TD Learning over Rate-Limited Channels}
\author{Sreejeet Maity and Aritra Mitra%
\thanks{The authors are with the Department of Electrical and Computer
Engineering, North Carolina State University, Raleigh. Email: {\tt\small \{\textcolor{winered}{smaity2},\textcolor{winered}{amitra2}\}@ncsu.edu}. This work is supported by the following grant from the National Science Foundation: NSF CAREER award 2542396.}} 
\begin{document}
\maketitle
\thispagestyle{empty}
\pagestyle{empty}
\begin{abstract}
Motivated by reinforcement learning subject to communication constraints, we study a policy evaluation problem with linear function approximation over a rate-limited channel. In our setup, an agent interacting with an unknown environment (modeled as a Markov Decision Process) transmits quantized estimates of a temporal-difference (TD) learning update direction to a server over a finite-capacity channel. Our goal is to investigate to what extent one can preserve the statistical convergence rates achievable when the channel has infinite capacity. To that end, we propose a new algorithm called Batched Adaptively Quantized TD (\textmd{\tt BAQ-TD}), which uses batching to reduce the variance of the transmitted information, and adaptive quantization to encode innovations. We prove a high-probability finite-time error bound for \textmd{\tt BAQ-TD}, and show that, above a certain finite channel capacity, \textmd{\tt BAQ-TD} preserves the convergence rate of unquantized TD up to logarithmic factors. Specifically, for a $d$-dimensional value function parameter, we establish that transmitting a total of $O(d \log(T))$ bits over a horizon of length $T$ suffices to achieve the order-optimal rate of $\tilde{O}(1/T)$, marking an exponential improvement in communication-complexity over prior work.
\end{abstract}
\section{Introduction}
\label{sec:intro}
The classical paradigm of networked control seeks to characterize how the presence of a communication network impacts closed-loop performance~\cite{tatikonda2004control, nair2007feedback, schenato2007foundations}. Specifically, when data collected from multiple sensors is transmitted over non-ideal channels that are rate-limited, noisy, lossy, and introduce latencies, the goal is to understand how tasks such as state estimation, stabilization, and optimal control design get affected. Although the study of such questions is quite mature by now, the central pervasive assumption in networked control is that \emph{the system model is perfectly known}. Realizing that such an assumption is impractical for modern, complex dynamical systems, over the last decade, there has been a gradual shift of interest towards data-driven control/reinforcement learning (RL) algorithms. Typically, the performance of RL algorithms is assessed via \emph{sample complexity}, i.e., the number of samples (or interactions with the unknown environment) needed to learn a near-optimal policy. Communication once again becomes a central consideration in this context when the sensing and decision-making modules are not co-located. Surprisingly, even for the simplest of RL problems, the impact of communication constraints on sample-complexity is not well understood. Motivated by this gap, in this paper, we aim to shed light on the interplay between control, communication, and model-uncertainty.  

\textbf{The Setting.} A natural starting point for our investigation is (arguably) the simplest RL problem of \emph{policy evaluation} - estimating the value function corresponding to a given policy for subsequent policy improvement. We consider such a problem in discrete Markov Decision Processes (MDPs) with a finite number of states and actions. To allow for large state-spaces, we follow the standard idea of linear function approximation (LFA), where the value function estimate is parameterized by a linear function class. The classical Temporal Difference (TD) learning algorithm~\cite{tsitsiklisroy} uses data generated by the policy (to be evaluated) to iteratively update the value-function parameter. While both asymptotic~\cite{tsitsiklisroy} and finite-sample~\cite{bhandari_finite, srikant, patil2023, mitra2024simple} convergence results for TD learning exist in the literature, our specific focus here is to characterize how communication constraints impact such convergence. 

To do so, we take inspiration from the networked control literature~\cite{tatikonda2004control}, and consider a setting where a worker agent interacts directly with the MDP, collects raw data samples, and communicates encoded information to a remote server over a noiseless, rate-limited up-link channel that can support only a finite-bit message per transmission; see Fig.~\ref{fig:system-model}. The agent uses observed data to maintain noisy estimates of the TD direction and transmit quantized versions of such estimates. The server uses a decoded version of the quantized TD direction to update the value function parameter which is transmitted back to the agent over a perfect down-link channel of infinite capacity. In our setup, one can essentially view the agent as the ``sensor" responsible for data collection and estimation, and the server as the ``decision-maker" responsible for parameter updates. 

Given this premise, we ask: \emph{With a 
time-horizon of $T$ time-steps (equivalently, a sample budget of $T$), how many bits of information need to be transmitted from the agent to the server to match (if at all possible) the statistical accuracy of unquantized TD learning, i.e., TD learning with no communication constraints?} The only two papers we know of that provide partial answers to this question are~\cite{mitra2023temporal} and~\cite{dal2024finite} which explore the use of generic error-feedback and randomized quantization mechanisms, respectively. In each of these papers, the total communication cost scales \emph{linearly} with the sample budget $T$, which can be prohibitively large for communication-constrained environments. Our main contribution in this paper is to offer a new variant of the $\textrm{TD(0)}$ algorithm where the corresponding communication complexity scales merely \emph{logarithmically} in $T$, \textbf{marking an exponential improvement} relative to the prior work in~\cite{mitra2023temporal, dal2024finite}. In what follows, we elaborate on this contribution. 

$\bullet$ \textbf{Algorithmic Contribution.} There are two sources of error in our setup which complicate the process of encoder-decoder design. First, the randomness in the data (rewards and state transitions) causes fluctuations in the empirically estimated TD directions. Such random fluctuations need to be accurately accounted for in the design of the encoding radius: if the radius is too small, then it may not be able to capture worst-case deviations around the mean estimate; if the radius is too large, then the resulting sample-complexity bounds might be too loose. Second, the rate-limited channel distorts the estimated TD direction, adding a second layer of uncertainty. The challenge lies in ensuring that the two errors alluded to above do not pile up across iterations. 

In response to this challenge, we develop \aqtd, a novel variant of the $\textrm{TD}(0)$ algorithm that relies on two key ideas: (i) \textbf{model-based variance reduction} to reduce the variance of each estimated TD direction, and (ii) \textbf{adaptive quantization} to encode innovations in the TD directions. Specifically, our algorithm runs in epochs, and in each epoch, the agent uses the data collected during that epoch to generate a ``high-accuracy" variance reduced estimate of the ideal steady-state $\textrm{TD}(0)$ direction. The simple intuition here is that reducing the variance leads to a lower uncertainty, which, in turn, translates to fewer bits needed to encode the uncertainty. To further save on bits, we only encode ``innovations" capturing what is new at the agent (based on fresh data) relative to what is already known at the server from previous transmissions. Designing an adaptive encoding radius that contains such a \emph{random} innovation process (with high probability) is the chief non-trivial task; we show how this can be done in Section~\ref{sec:algorithm}. 

$\bullet$ \textbf{Theoretical contribution.} Given $T$ samples, in Theorem~\ref{thm:stochastic}, we provide a high-probability convergence rate on the order of $\widetilde{\mathcal{O}}(1/T)$ for \aqtd. Such a rate is consistent with what is known for unquantized $\textrm{TD}(0)$~\cite{bhandari_finite, srikant, patil2023, mitra2024simple}, and is order-optimal in the sample-size $T$. Thus, we show that statistical rates can be preserved despite finite channel capacities. Interestingly, the channel capacity needed to achieve this result does not depend on $T$, but rather scales as $d \log(d/\omega)$, where $d$ is the dimension of the value function parameter, and $\omega$ is a problem-dependent constant that we specify in Section~\ref{sec:proof}. Moreover, we show that $\widetilde{\mathcal{O}}(\log(T))$ transmissions suffice to preserve order-optimal rates, leading to a total communication cost on the order $\widetilde{\mathcal{O}}(d\log(T))$. 

\textbf{Related Work.} Recently, a growing body of work has started looking at the effect of communication constraints for decision-making problems under uncertainty, with stochastic bandits~\cite{hanna2022solving, salgia2023distributed, mitra2023linear} and tabular MDPs~\cite{salgia2024sample} as natural starting points. Beyond the simple tabular case, the interplay between function approximation and bit constraints has been explored in~\cite{mitra2023temporal, dal2024finite}. Moreover, for continuous state-action spaces, some recent papers conducted similar studies on policy gradient algorithms for the linear quadratic regulator (LQR)~\cite{yang2024over, mitra24a}. In fact, we build on the adaptive quantization idea in~\cite{mitra24a}. However, all of the papers~\cite{mitra2023temporal, dal2024finite, yang2024over, mitra24a} incur communication costs that are linear in sample-size/time-horizon. In contrast, we show that despite errors resulting from function approximation, random data, and channel-induced distortions, logarithmic-in-sample size communications costs are adequate for preserving ``ideal" convergence rates. 
\section{Background and Problem Formulation}
\label{sec:model}
We consider a discounted Markov decision process
$\mc{M}=(\mc{S},\mc{A},\mc{P},\mc{R},\gamma)$, where $\mc{S}$
and $\mc{A}$ are the (finite) state and action spaces, $\mc{P}$ is the
collection of transition kernels, $\mc{R}$ specifies reward
functions, and $\gamma\in[0,1)$ is the discount factor. We consider deterministic policies, where each deterministic policy $\mu: \mc{S} \rightarrow \mc{A}$ maps a state to an action. A fixed
policy $\mu$ induces a Markov reward process with a transition matrix
$P_\mu$, and a reward function $R_{\mu}: \mc{S} \to \mathbb{R}$. For each $s\in\mc{S}$, let $r(s)$ denote the noisy random one-step
reward with distribution $\mc{D}_\mu(\cdot\mid s)$ and mean
$R_\mu(s)=\E[r(s)]$.
We assume\footnote{The sub-Gaussianity assumption simplifies the exposition;
bounded-variance heavy-tailed rewards can be handled via
tools used in~\cite{maity2025,pmlr-v306-maity26a}.}
that $|R_\mu(s)|\le\bar r$ and that the centered reward
$r(s)-R_\mu(s)$ is sub-Gaussian with variance proxy
$\sigma^2$, i.e.,
$\E\big[\exp(\lambda(r(s)-R_\mu(s)))\big]
\le\exp(\lambda^2\sigma^2/2)$
for every $s\in\mc{S}$ and $\lambda\in\R$.
The long-term return obtained from playing policy $\mu$ in the MDP $\mc{M}$, starting from an initial state $s$, is given by the value function $V_{\mu}(s)$, defined formally as
\begin{equation}\label{eqn:v_cum}
 V_\mu(s)
 =
 \E\!\left[
   \sum_{t=0}^{\infty}\gamma^t R_\mu(s_t)
   \,\middle|\,s_0=s,\mu
 \right],
\end{equation}
where $s_t$ represents the state of the Markov chain (induced by $\mu$)
at time $t$, with $s_0=s$. To initiate the study of communication-constrained RL, we consider one of the simplest RL problems in this paper, namely \emph{policy evaluation} - the
task of evaluating $V_{\mu}$ corresponding to a given policy
$\mu$, when
$R_{\mu}=[R_{\mu}(s)]_{s\in\mc{S}}\in\R^{|\mc{S}|}$
and $P_{\mu}$ are \emph{unknown}. We make the following standard assumption~\cite{tsitsiklisroy, bhandari_finite, srikant, mitra2024simple}. 
\begin{assumption}\label{ass:ergodic}
The Markov chain ${s_t}$ induced by the policy $\mu$
is aperiodic and irreducible.
\end{assumption}
Assumption~\ref{ass:ergodic} ensures
that the Markov chain induced by $\mu$ admits a unique
stationary distribution $\nu$ with full support. To isolate the challenges introduced by communication constraints, we adopt the commonly used i.i.d.\ stationary sampling model~\cite{korda, dalal, narayanan, bhandari_finite}:
at each time-step $t$, the agent receives an independent
transition sample $(s_t,r_t,s'_t)$, where $s_t\sim\nu$, $s'_t \sim P_{\mu}(s_t, \cdot)$, and $r_t \sim \mc{D}_{\mu}(\cdot| s_t).$ The results in this paper can be extended to a single-trajectory Markovian setting using sub-sampling and blocking arguments~\cite{dorfman, maity2025}. 

\textbf{Linear Function Approximation (LFA).} To accommodate scalability to large state spaces, one typically considers a parametric approximation $\hat{V}_\theta$ of $V_\mu$. The simplest form of such an approximation is when $\hat{V}_\theta$ belongs to the linear subspace spanned by a set $\{\phi_k\}_{k\in [d]}$ of $d \ll |\mc{S}|:=p$  basis vectors, where $\phi_k =[\phi_k(1), \ldots, \phi_k(p)]^{\top} \in \mathbb{R}^{p}$. Let $\Phi \in \mathbb{R}^{p \times d}$ be a matrix with $\phi_k$ as its $k$-th column, and its $s$-th row $\phi(s) \in \mathbb{R}^{d}$ representing the feature vector for state $s$. We then have $\hat{V}_\theta=\Phi \theta$, and $\hat V_\theta(s)=\sum_{k=1}^d\theta_k\phi_k(s)
=\phi(s)^\top\theta$, where $\theta=[\theta_1,\ldots,\theta_d]^\top\in\R^d$ is the parametric approximation vector. For our analysis, we make the standard assumption~\cite{bhandari_finite, srikant, mitra2024simple} that the columns of $\Phi$ are linearly independent, and that the feature vectors are normalized, i.e., for each $s \in \mc{S}$, $\Vert \phi(s) \Vert \leq 1$.\footnote{
Unless otherwise stated, vector norms are Euclidean
and matrix norms are the corresponding induced spectral norms.} Given the above setup, the goal is to estimate the ``best-fit" parameter $\theta^*$ that minimizes the distance between $\hat{V}_{\theta}$ and $V_{\mu}$ with respect to (w.r.t.) a suitable norm. 

\begin{figure}[t]
\begin{center}
\includegraphics[scale=0.36]{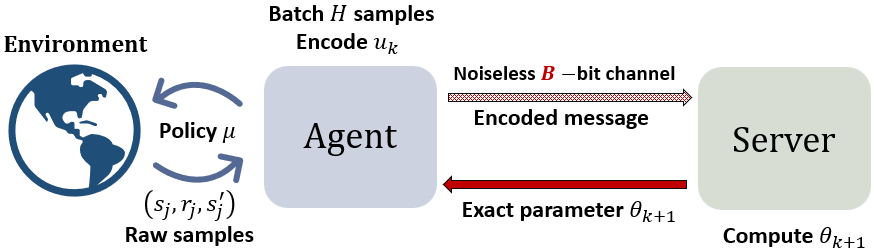}
\vspace{-5mm}
\end{center}
\caption{Agent-server architecture for
\textcolor{winered}{\texttt{BAQ-TD}}, presented in
Algorithm~\ref{alg:baqtd}. During each epoch $k$, the agent uses $H$ transition samples from the unknown MDP under policy $\mu$ to
estimate the TD direction and transmits the encoded
innovation over a noiseless,
rate-limited channel.
The server reconstructs the direction, updates its
parameter to $\theta_{k+1}$, and returns this parameter
exactly over an unconstrained noiseless reverse channel.}
\label{fig:system-model}
\end{figure}

\textbf{The $\mathrm{TD}(0)$ Algorithm with LFA.} For achieving this goal, we now introduce the classical $\mathrm{TD}(0)$ algorithm~\cite{sutton1988} which processes observation tuples $\{O_t=(s_t, r_t, s'_t)\}$ generated by the policy $\mu$ to maintain a sequence of estimates $\{\theta_t\}$ of the best-fit parameter $\theta^*$. To see how this is done, for any $\theta\in\R^d$, define the
$\mathrm{TD}(0)$ direction:
\(
g_t(\theta)
=
[
  r_t+\gamma\phi(s'_t)^\top\theta
  -\phi(s_t)^\top\theta
]\phi(s_t).
\)
The standard $\mathrm{TD}(0)$ update is then given by
\begin{equation}
\theta_{t+1}=\theta_t+\alpha_t g_t(\theta_t),
\label{eqn: TD(0)}
\end{equation}
where $\{\alpha_t\} \in(0,1)$ is a suitable step-size sequence. Under our assumption of i.i.d. sampling from the stationary distribution $\nu$, the steady-state ``mean-path" $\mathrm{TD}(0)$ direction can be defined as: $\gbar(\theta):= \E[g_t(\theta)] = b-A\theta$, where 
\(
A:=\E\!\left[
  \phi(s_t)\bigl(\phi(s_t)-\gamma\phi(s'_t)\bigr)^\top
\right]\) and \(
b:=\E[r_t\phi(s_t)]. \) Under Assumption~\ref{ass:ergodic}, and the assumptions we made on the feature matrix $\Phi$, it turns out that $A$ is invertible. Accordingly, define $\theta^\star=A^{-1}b$. In the seminal paper~\cite{tsitsiklisroy}, the authors showed that under mild conditions, the iterate sequence $\{\theta_t\}$ generated by~\eqref{eqn: TD(0)} converges asymptotically almost surely to $\theta^*$, which is the unique fixed point of a certain projected Bellman equation. More recent work has established non-asymptotic finite-time convergence rates, proving that after $T$ iterations of~\eqref{eqn: TD(0)}, the squared error $\Vert \theta_T - \theta^*\Vert^2$ is on the order $\tilde{\mc{O}}(1/T)$ in expectation~\cite{bhandari_finite, srikant, mitra2024simple}. We investigate whether similar rates can be preserved subject to the communication constraints that we describe next. 

\textbf{Rate-Limited $\mathrm{TD}(0)$ Policy Evaluation.} We study estimation of the TD(0) fixed point
$\theta^\star$ in the agent--server architecture
illustrated in Figure~\ref{fig:system-model}. In this setup, an agent interacting directly with the MDP $\mc{M}$ observes data tuples of the form $\{O_t=(s_t, r_t, s'_t)\}$ and transmits encoded information to a server over a finite-capacity channel that supports noiseless and error-free transmission of $B$ bits per channel use. Complying with standard communication protocols in distributed/federated RL, we impose an additional information constraint that the agent can only transmit encoded estimates of the TD(0) update direction, but not the raw data tuples. The channel from the server to the agent is assumed to be lossless with infinite capacity. Given this communication model, our \textbf{goal} is to design (i) an encoding scheme at the agent, and (ii) a parameter update rule at the server that continues to guarantee convergence to $\theta^*$ (if possible). Furthermore, we seek to quantify the interplay between channel capacity and finite-time convergence rates. 

\begin{tcolorbox}[colback=blue!5!white,colframe=blue,breakable]
\begin{problem}\label{prob:main}
Given $T$ i.i.d.\ stationary transition samples
$\{(s_t,r_t,s'_t)\}_{t=0}^{T-1}$ and a failure probability
$\delta\in(0,1)$, construct an estimate
$\hat\theta$ of $\theta^\star$ using fixed-length $B$-bit
agent-to-server messages, and establish a bound of the form: $\Vert \hat{\theta} - \theta^* \Vert^2 \leq e(\delta, T)$ that holds with probability at least $1-\delta$. Investigate whether with a suitably large capacity $B$, it is possible to preserve the order-optimal rate of $e(\delta,T) = \tilde{\mc{O}}(\log(1/\delta)/T).$ 
\end{problem}
\end{tcolorbox}

\textbf{Challenges.} Even for the standard $\mathrm{TD}(0)$ algorithm with LFA, it is non-trivial to obtain high-probability concentration bounds due to the randomness associated with the noisy TD update directions. In our setup, the channel from the agent to the server injects additional uncertainty by distorting the estimates generated by the agent. Thus, we need to contend with two different forms of uncertainty: one due to sampling, and the other due to finite-bit quantization. Nonetheless, in Section~\ref{sec:algorithm}, we develop an algorithm that manages to preserve the order-optimal rate of $\tilde{\mc{O}}(1/T)$, using a total of just $\tilde{\mc{O}}(d \log(T))$ bits over a time-horizon of length $T$. 

\section{The Proposed Algorithm}
\vspace{-1mm}
\label{sec:algorithm}
\noindent{\textbf{Midpoint Encoding}.} Before presenting our main approach outlined in Algorithm~\ref{alg:baqtd}, we specify a key finite-bit encoding subroutine. Say we wish to encode a $d$-dimensional vector $u$ using $m$ bits per coordinate, so that each message contains $B=dm$ bits. The encoder then takes as input the vector
$u\in\R^d$, a radius $\rho>0$ which determines the encoder's input range, and an 
integer precision $m\ge1$. The encoder and decoder share
$\rho$, $d$, and $m$. To define the encoding operation for every $u$, including
when $\norm{u}>\rho$, we first clip $u$ to radius $\rho$: $ \sat_\rho(u):= u$ if $\Vert u \Vert \leq \rho$; else, $\sat_\rho(u):= \rho u/\Vert u \Vert.$ 
With $v=\sat_\rho(u)$, clearly $\norm{v}\le\rho$, and
hence, $v_i \in [-\rho,\rho], \forall i \in [d]$, where $v_i$ is the $i$-th coordinate of $v$. This informs a simple midpoint encoding scheme in which 
we partition the interval $[-\rho,\rho]$ into $2^m$ bins of equal width. For each coordinate $v_i \in [-\rho,\rho]$, let $\tilde{v}_i$ be the center of the bin containing $v_i$. Then, the quantized version of a vector $u \in  \mathbb{R}^d$ can be succinctly denoted by $Q_{m,\rho}(v) := \begin{bmatrix} \tilde{v}_1 & \tilde{v}_2 & \cdots & \tilde{v}_d \end{bmatrix}^T, $ where $v= \sat_\rho(u).$ It should be clear that this midpoint encoding scheme requires precisely $m$-bits per coordinate. The following simple result provides an useful bound on the quantization error.

\begin{lemma}\textbf{(Quantization Error)}
\label{lem:quant-error}
Suppose $v\in\R^d$ satisfies
$\norm{v}\le\rho$. Then
\begin{equation}\label{eq:quant-error}
\norm{Q_{m,\rho}(v)-v}\le\eta\rho,
\quad
\eta=\frac{\sqrt d}{2^m}.
\end{equation}
\end{lemma}

\begin{proof} The proof is standard and we only provide it here for completeness. Since $\norm{v}\le\rho$, each  $v_i$ lies
in $[-\rho,\rho]$. From the description of the midpoint encoder, recall that the $i$-th component of $Q_{m,\rho}(v)$, namely $\tilde{v}_i$ is the quantized version of $v_i$, and by construction, $\vert v_i - \tilde{v}_i \vert \leq \rho/2^m.$ 
Thus,
$\norm{Q_{m,\rho}(v)-v}^2
=\sum_{i=1}^d |\tilde{v}_i-v_i|^2
\le d\rho^2/2^{2m}$, from which, the claimed bound follows trivially.
\end{proof}

In particular, if $\norm{u}\le\rho$, then
$\sat_\rho(u)=u$, and \eqref{eq:quant-error} yields 
$\norm{Q_{m,\rho}(\sat_\rho(u))-u}\le\eta\rho$. In what follows, we explain how our proposed \texttt{BAQ-TD} algorithm leverages the midpoint encoding subroutine described above. 

\noindent \textbf{Batched Adaptively Quantized TD}: We now introduce our algorithm, \texttt{BAQ-TD}, which employs the principle of adaptive quantization to encode ``innovations" in the $\mathrm{TD}(0)$ update direction. For our setting, this poses a challenge since such innovations inherit randomness from the sampling-induced noise in the update directions. Intuitively, by reducing the uncertainty of the innovation process, one can reduce the number of bits needed to encode the innovations with a given precision. To that end, our main idea is to maintain \emph{variance-reduced} estimates of the $\mathrm{TD}(0)$ update direction at the agent by batching the $T$ samples collected online into epochs of suitably large length. In what follows, we provide the details of our two core algorithmic ideas: batched estimation and adaptive quantization.  

$\bullet$~\textbf{Batched TD-direction estimation.}
We partition the horizon/sample budget $T$ into $K$ epochs of $H$ samples each, with $T=KH$. For epoch $k=0,\ldots,K-1$, define its sample indices by
$\mathcal I_k:=\{kH,\ldots,(k+1)H-1\}$. At the beginning of each epoch $k$, the server sends its current parameter $\theta_k\in\R^d$,
previous decoded direction $h_{k-1}$, and radius $\rho_k$
to the agent. We will soon explain how $\theta_k$, $h_{k-1}$ and $\rho_k$ are initialized and updated in every epoch. The agent then collects $H$ observation tuples
$\{(s_j,r_j,s'_j)\}_{j\in\mathcal I_k}$ sequentially based on the i.i.d. sampling model in Section~\ref{sec:model}. Using these samples and the received parameter $\theta_k$,
the agent computes an estimate $Y_k$ of
the mean-path $\mathrm{TD(0)}$ direction $\bar{g}(\theta_k)$ as follows:
\begin{equation}\label{eq:Y}
Y_k
=
\frac1H\sum_{j\in\mathcal I_k}\phi(s_j)
\left[
r_j+\gamma\phi(s'_j)^\top\theta_k
-\phi(s_j)^\top\theta_k
\right].
\end{equation}
Given that each of the objects being averaged above feature a common ``frozen" parameter $\theta_k$, it is not hard to see that under our i.i.d. sampling model, $Y_k$ is an unbiased estimate of $\bar{g}(\theta_k)$ with variance scaling inversely with the batch size $H$. As we shall soon see, by controlling $H$, one can in turn control the uncertainty of the innovation process $\{u_k:= Y_k - h_{k-1}\}$, and the radius $\rho_k$ of the ball needed to contain such a process (with high probability). 

\begin{algorithm}[t]
\caption{Batched Adaptively Quantized TD (\texttt{BAQ-TD})}
\label{alg:baqtd}
\begin{algorithmic}[1]
\State \textbf{Input:} $T,K,\delta,\alpha,m,\theta_0$
satisfying the conditions of Theorem~\ref{thm:stochastic}.
\State \textbf{Initialize:} Set $H=T/K$ and
$\eta=\sqrt d/2^m$; compute $C_\delta$ by
\eqref{eq:Cdelta}; set $h_{-1},\rho_0$ as per \eqref{eq:init}.
\For{$k=0,\ldots,K-1$}
    \State
    (\(\bullet\)) \textbf{Server (\textcolor{mygreen}{\tt{Downlink}}):} 
    \Statex
    $\textcolor{mygreen}{\blacktriangleright}$
    Sends $(\theta_k,h_{k-1},\rho_k)$ to the agent.

    \State $(\bullet)$
    \textbf{Agent}
    (\textcolor{myblue}{\texttt{Compute and transmit}}):
    \Statex
    $\textcolor{myblue}{\blacktriangleright}$
    Draws i.i.d.\ tuples
    $\{(s_j,r_j,s'_j)\}_{j\in\mathcal I_k}$,
    where $|\mathcal I_k|=H$.
    \Statex
    $\textcolor{myblue}{\blacktriangleright}$
    Computes $Y_k$ using \eqref{eq:Y}.
    \Statex $\textcolor{myblue}{\blacktriangleright}$
Sets $u_k=Y_k-h_{k-1}$ and $v_k=\sat_{\rho_k}(u_k)$.
\Statex $\textcolor{myblue}{\blacktriangleright}$
Sends a $dm$-bit message encoding the innovation $z_k=Q_{m,\rho_k}(v_k)$ using a scalar midpoint encoder.

    \State $(\bullet)$
    \textbf{Server}
    (\textcolor{mygreen}{\texttt{Decode and update}}):
    \Statex
    $\textcolor{mygreen}{\blacktriangleright}$
    Decodes $z_k=Q_{m,\rho_k}(v_k)$.
    \Statex
    $\textcolor{mygreen}{\blacktriangleright}$
    Sets $h_k=h_{k-1}+z_k$ and
    $\theta_{k+1}=\theta_k+\alpha h_k$.
    \Statex
    $\textcolor{mygreen}{\blacktriangleright}$
    Computes $\rho_{k+1}$ by \eqref{eq:range-update}.
\EndFor
\State \textbf{Return} $\theta_K$.
\end{algorithmic}
\end{algorithm}

$\bullet$~\textbf{Encoding and Decoding the Innovation.} The agent forms the innovation $u_k:=Y_k-h_{k-1}$,
which measures the difference between the current
estimated direction at the agent and the previously decoded direction
at the server. Using the scalar midpoint encoder described earlier, the agent transmits a $dm$-bit message corresponding to the quantized signal $z_k:= Q_{m,\rho_k}(v_k),$ where $v_k = \sat_{\rho_k}(u_k)$. The server decodes the message to recover $z_k$ perfectly, and then uses it to reconstruct the $\mathrm{TD}(0)$ direction and update the model parameter as follows: 
\begin{equation}\label{eq:h-update}
h_k=h_{k-1}+z_k, \quad \theta_{k+1}=\theta_k+\alpha h_k,
\end{equation}
where $\alpha$ is a step-size that we will specify in the next section. It remains to explain how the radius $\rho_k$ is adaptively tuned in each epoch to contain the innovation process. To do so, given $K,H,m\ge1$, a confidence level $\delta\in(0,1)$, and statistics $\bar r$ and $\sigma$ of the reward distribution (defined
in Section~\ref{sec:model}), set
\begin{equation}\label{eq:Cdelta}
C_\delta^2
=2\max\!\left\{4d^2,\ d(\bar r^2+\sigma^2)\right\}
  \log\!\left(\frac{4Kd^2}{\delta}\right),
\end{equation}
and define the fixed batch-error allowance
$\epsilon_H:=C_\delta/\sqrt H$; note that $\epsilon_H$ can be computed offline. The agent and server share these quantities and the
initial parameter $\theta_0$. We initialize the decoded direction and encoding radius as
\begin{equation}\label{eq:init}
h_{-1}=0,
\quad
\rho_0
=
\bar r+2\norm{\theta_0}
+\epsilon_H(\norm{\theta_0}+1).
\end{equation}

Let $\eta=\sqrt d/2^m$ be the quantization constant from
Lemma~\ref{lem:quant-error}. After updating the parameter, the server adjusts the encoding radius as follows: 
\begin{equation}\label{eq:range-update}
\rho_{k+1}
=
\eta\rho_k+2\alpha\norm{h_k}
+2\epsilon_H(\norm{\theta_k}+1).
\end{equation}

The next server transmission supplies $(\theta_{k+1},h_{k},\rho_{k+1})$ to the
agent, ensuring that both parties use the same
reference for the encoding-decoding process. This also enables the server to decode the $\{z_k\}$ sequence \emph{exactly} from the received $B=dm$-bit agent messages. Let us now briefly explain the rationale behind the adaptive range-update rule in~\eqref{eq:range-update}

$\bullet$~\textbf{Adaptive encoding radius.} The radius $\rho_k$ is chosen to contain the next innovation
$Y_{k+1}-h_k$. The first term on the R.H.S. of~\eqref{eq:range-update}, namely $\eta\rho_k$, is informed by Lemma~\ref{lem:quant-error}, and bounds the reconstruction
error in the current decoded direction whenever
saturation is inactive, i.e., when $\Vert u_k \Vert \leq \rho_k$. Since $Y_{k+1}$ is computed at $\theta_{k+1}$ and $h_k$ corresponds to $\theta_k$, the gap $Y_{k+1}-h_k$ depends on the parameter change
$\theta_{k+1}-\theta_k=\alpha h_k$, explaining the second term on the R.H.S. of \eqref{eq:range-update}. Finally, the encoding radius needs to account for the errors in the batched $\textrm{TD}(0)$ estimation process.  In this regard, we show later in Lemma~\ref{lem:concentration} that,
with probability at least $1-\delta$, the empirical
TD matrices and reward vectors are simultaneously
within $\epsilon_H$ of $A$ and $b$, respectively,
across all $K$ batches. As a result, the precision level $\epsilon_H$ shows up in the last term of~\eqref{eq:range-update}. 
Together with the initial radius $\rho_0$, these bounds
ensure inductively that every transmitted innovation
satisfies $\norm{u_k}\le\rho_k$ with high-probability; see Lemma~\ref{lem:no-saturation}. 

Our method processes $H$ samples per communication
round, yielding $K$ parameter updates and $K$ fixed-length
messages over the full sample budget.
Thus, over the time-horizon of length $T$, the total communication cost is $Kdm$ bits. In the next section, we show that for a suitably chosen parameter $m$, $K = {\mc{O}}(\log T)$ epochs suffice to obtain the desired $\tilde{\mc{O}}(1/T)$ rate. The steps of our algorithm are outlined in Algorithm~\ref{alg:baqtd}, where 
$\textcolor{myblue}{\blacktriangleright}$ and
$\textcolor{mygreen}{\blacktriangleright}$ denote operations
performed by the agent and the server, respectively. 
\section{Main Results}
\label{sec:proof}
We now establish a finite-time convergence guarantee
for \texttt{BAQ-TD}.
To do so, let $e_k:=\theta_k-\theta^\star$ denote the parameter
error, and define a coupled Lyapunov function 
\(
V_k:=\norm{e_k}^2+\alpha\rho_k^2
\)
that jointly tracks the parameter error
and the encoding radius. Let $D$ be a diagonal matrix containing the entries of the stationary distribution $\nu$ along the diagonal, and define $\Sigma = \Phi^{\top} D \Phi$. Under Assumption~\ref{ass:ergodic} and the conditions we imposed on $\Phi$, one can verify that $\Sigma$ is positive definite with smallest eigenvalue $\tilde{\omega} \in (0,1)$. Defining $\omega:= \tilde{\omega}(1-\gamma)$, and recalling $C_{\delta}$ from~\eqref{eq:Cdelta}, we can state our main result. 

\begin{theorem}\textbf{(Finite-Time Convergence)}
\label{thm:stochastic}
Suppose Assumption~\ref{ass:ergodic} holds. 
Given $\delta \in (0,1)$, run Algorithm~\ref{alg:baqtd} with 
 stepsize
$\alpha\in(0,\omega/32]$ and precision satisfying
\(
m\ge
\left\lceil
\frac12\log_2\!\left(\frac{64d}{\omega}\right)
\right\rceil.
\)
Choose
$K=\lceil c_1\log T/(\alpha\omega)\rceil$
and suppose that the total sample count $T=KH$
satisfies $T\ge c_2KC_\delta^2/\omega^2$.
Then, with probability (w.p.) at least $1-\delta$,
the bound
\begin{equation}\label{eq:main-bound}
\norm{\theta_k-\theta^\star}^2
\le
(1-\alpha\omega)^kV_0
+
{\mc O}\!\left(
\frac{
C_\delta^2(1+\norm{\theta^\star})^2\log T
}{
\alpha\omega^3T
}
\right)
\end{equation}
holds simultaneously for all $k \in [K]$. Here, 
$V_0=\norm{\theta_0-\theta^\star}^2+\alpha\rho_0^2$
and $c_1,c_2$ are sufficiently large universal constants.
\end{theorem}

Each epoch uses $H$ samples and produces one parameter
update and one $dm$-bit agent-to-server message.
The total costs are therefore $KH$ samples,
$K$ messages, and $Kdm$ transmitted bits.
The following corollary, which follows directly from Theorem~\ref{thm:stochastic}, combines these communication
costs with the statistical guarantee.

\begin{corollary}\textbf{(Accuracy and Communication)}
\label{cor:communication}
Under the conditions of Theorem~\ref{thm:stochastic},
choose $m=\left\lceil\frac12\log_2\!\left(64d/\omega\right)\right\rceil$ and $\alpha=\omega/32$. 
Over the $T=KH$ time horizon, \texttt{BAQ-TD} transmits
$K=\lceil 32 c_1\log T/(\omega^2)\rceil$
fixed-length messages, each containing $dm$ bits.
The total number of bits transmitted is thus
\begin{equation}\label{eq:communication-bits}
Kdm
=
\mathcal{O}\!\left(
\frac{d\log(d/\omega)}{\omega^2}\log T
\right).
\end{equation}
For any $\delta\in(0,1)$, if $T$ satisfies the sample-size
condition of Theorem~\ref{thm:stochastic}, then with probability
at least $1-\delta$,
\begin{equation}\label{eqn:stat_rate}
\norm{\theta_K-\theta^\star}^2
\le \frac{V_0}{T}
+\frac{C_p\log T}{T}
 \log\!\left(\frac{d^2\log T}{\omega^2\delta}\right),
\end{equation}
where $C_p=Cc_1^2(1+\norm{\theta^\star})^2
\max\{4d^2,d(\bar r^2+\sigma^2)\}/\omega^4$, and $C>0$ is a  sufficiently large universal constant 
\end{corollary}

\noindent\textbf{Discussion of
Theorem~\ref{thm:stochastic} and Corollary~\ref{cor:communication}.}
Theorem~\ref{thm:stochastic} provides a finite-time
high-probability guarantee for TD learning with finite-bit communication.
The first term in~\eqref{eq:main-bound} is a geometrically decaying
transient term of the form  $e^{-\alpha\omega  k} V_0$, where $V_0=\norm{e_0}^2+\alpha\rho_0^2$ reflects the coupling between the parameter estimation error and errors due to quantization (captured by the encoding radius $\rho$). The second term in~\eqref{eq:main-bound} is a statistical error term that shrinks with the sample-size $T$. We discuss the sample and communication complexities of our method below. 

$\bullet$~\textbf{Statistical accuracy.}
The statistical residual obtained in our analysis is
of order
$C_\delta^2(1+\norm{\theta^\star})^2/(\omega^2H)$, where recall that $H$ is the size of each epoch. Clearly, increasing $H$ reduces the residual, but comes at the expense of fewer contractive epochs $K$ to control the transient (since $T=KH$). Fortunately, since the transient term decays exponentially in $K$, setting $K=\mc{O}(\log(T))$ suffices to achieve an overall rate of $\tilde{\mc{O}}(1/T)$, as established in Corollary~\ref{cor:communication}. Such a rate is consistent with prior non-asymptotic mean-squared error~\cite{bhandari_finite, srikant, mitra2024simple} and high-probability bounds~\cite{patil2023} for TD(0), and is order-optimal in its dependence on the sample size $T$.  

$\bullet$~\textbf{Communication Costs.} In terms of communication complexity, our results provide two key messages. First, we show that with $m=\mc{O}(\log(d/\omega))$, a fixed channel capacity of $B=dm$ bits suffices to preserve the ideal rate of $\tilde{\mc{O}}(1/T)$, i.e., we establish that above a certain finite channel capacity, it is possible to ensure no loss in performance. Notably, the requirement on $m$ that we identify depends only on the feature-dimension $d$ and ``strong-convexity-like" parameter $\omega$,
but not on the batch size $H$ or total budget $T$. Thus, even when the sample budget $T$ increases, the number of bits per message can continue to remain fixed. 

While this settles the question of how much information needs to be sent \emph{per round}, it leaves open the issue of determining how frequently the agent should communicate with the server. Since the number of communication rounds is precisely $K$, and $K=\log(T)$, we show that the number of such rounds exhibits a modest logarithmic scaling with the horizon $T$. In sharp contrast, the compressed TD schemes
in~\cite{mitra2023temporal,dal2024finite}
communicate at each time-step, incurring a total communication cost that scales linearly with $T$. Thus, we achieve an \emph{exponential improvement} in communication costs relative to prior work, and provide the first results to show that one can continue to obtain high-probability rates on the order $\tilde{\mc{O}}(\log(1/\delta)/T)$ with a total of just $\mc{O}(d \log(T))$ bits. 

\begin{figure*}[t]
\centering
\begin{tabular}{@{}ccc@{}}
\hspace{-4mm}\includegraphics[scale=0.25]{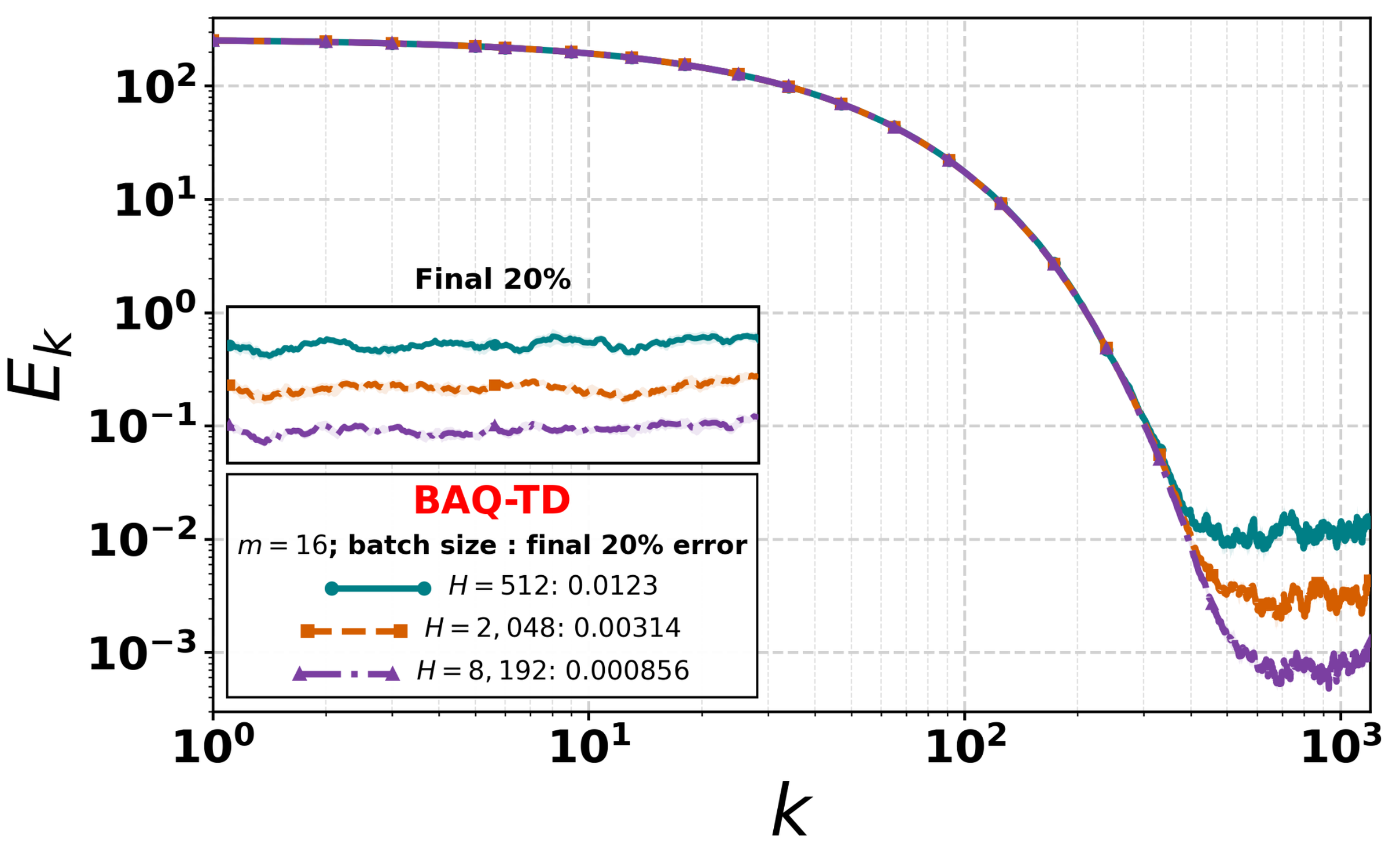}
&
\hspace{-4mm}\includegraphics[scale=0.25]{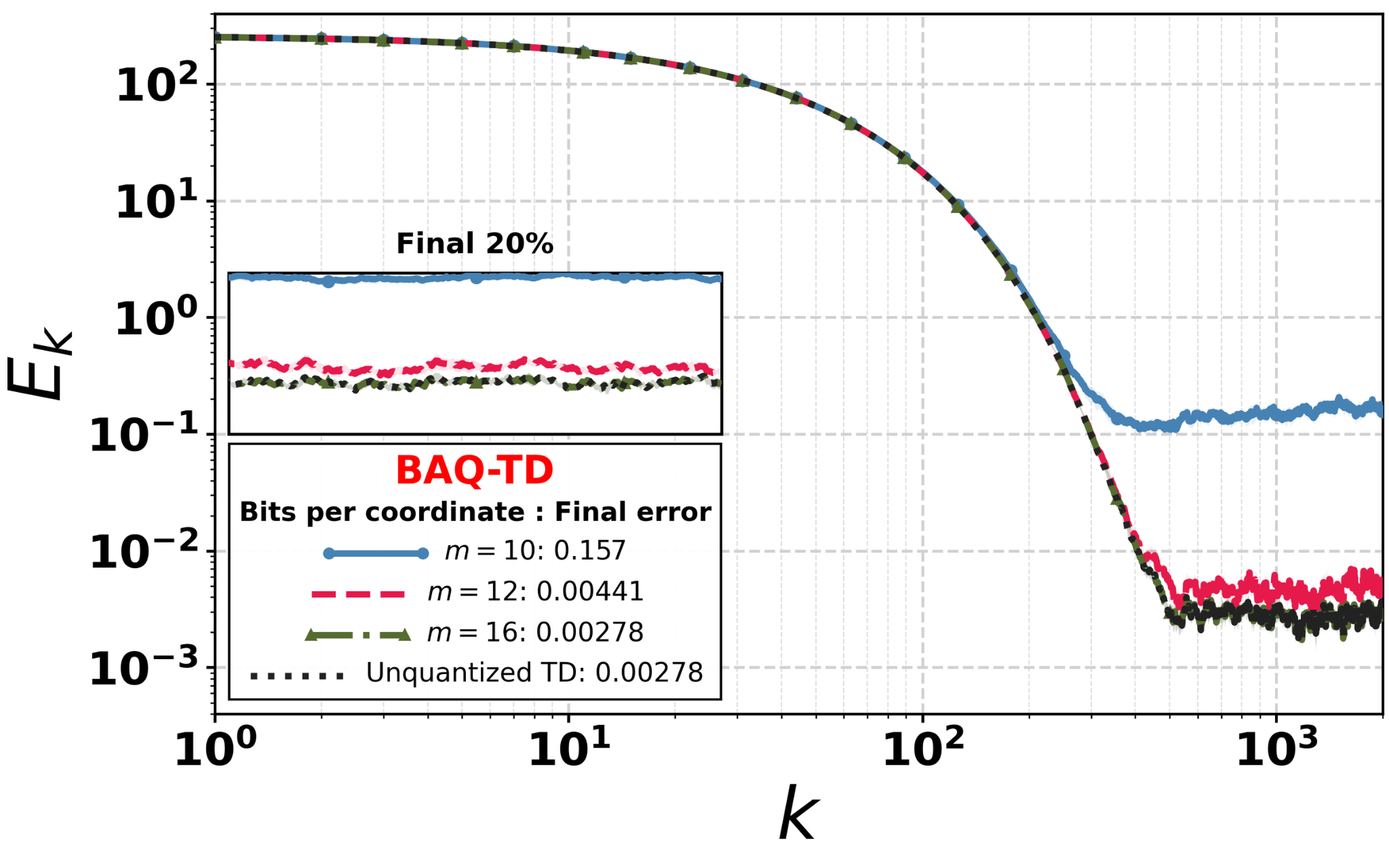}
&
\hspace{-4mm}\includegraphics[scale=0.06]{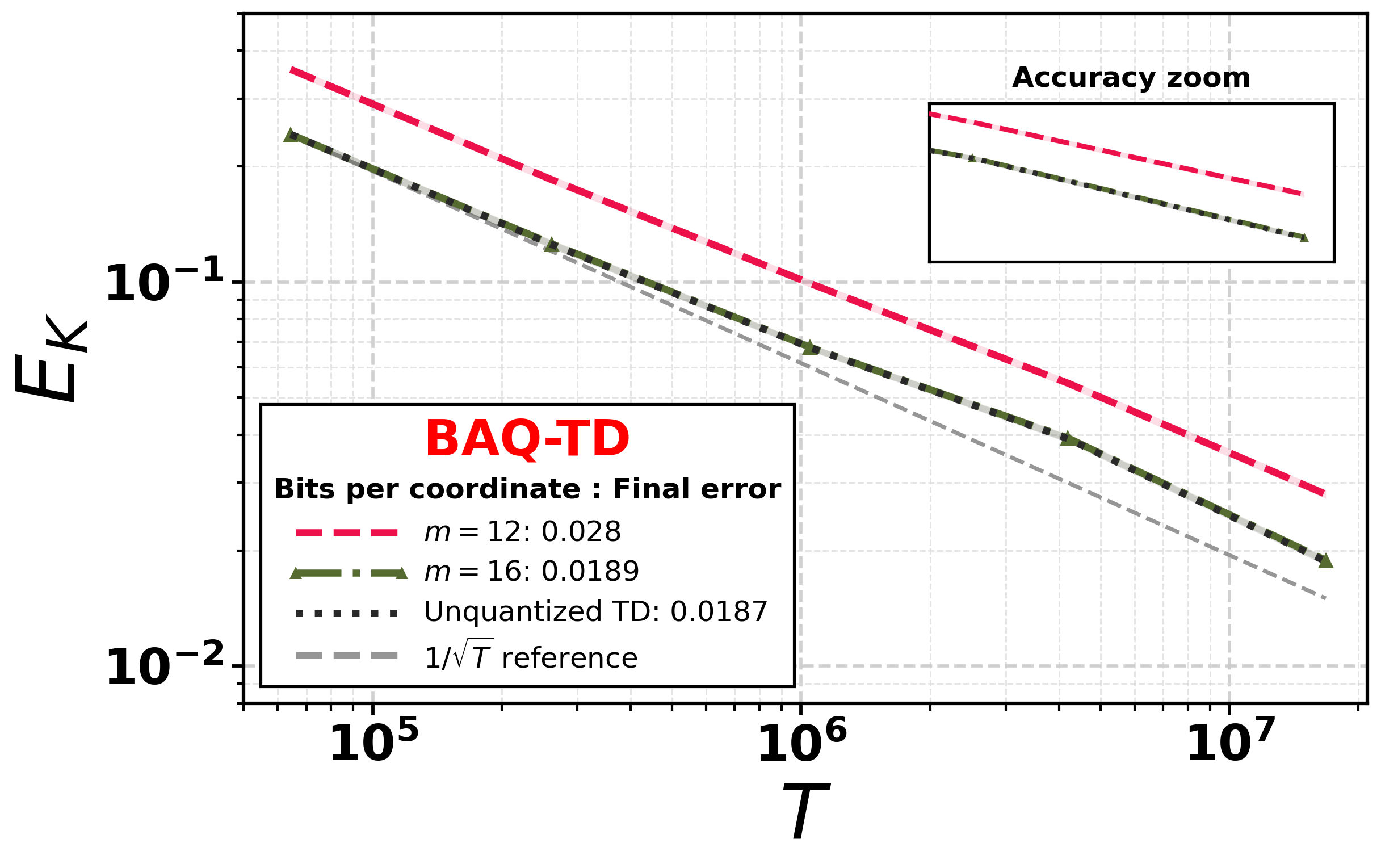}
\end{tabular}
\caption{\textbf{(Left)} Evolution of the estimation error
$E_k=\norm{\theta_k-\theta^\star}_2$ for
\texttt{BAQ-TD} with batch sizes
$H\in\{512,2048,8192\}$, precision $m=16$,
and $K=1000$ epochs.
\textbf{(Middle)} Results for precisions
$m\in\{10,12,16\}$, compared with unquantized TD. \textbf{(Right)} Terminal estimation error
$E_K=\|\theta_K-\theta^\star\|_2$ versus sample budget $T$
for \texttt{BAQ-TD} with $m\in\{12,16\}$
and unquantized TD. The dashed line shows the $T^{-1/2}$ scaling suggested by our square-error bound.
}
\label{fig:sim-1}
\end{figure*}
\section{Simulation Results}
\label{sec:simulation}
We evaluate the performance of \texttt{BAQ-TD} on a finite discounted Markov decision process with 100 states and discount factor $\gamma=0.5$. The mean reward at each state is bounded above by $\bar r\le 5$, and the reward-noise variance satisfies $\sigma^2\le 1$. The value function is approximated using $d=10$ features, with each feature vector normalized to have norm at most one. We compute the stationary distribution and the corresponding TD fixed point $\theta^\star$ using the known model. Each plot in Figure~\ref{fig:sim-1} reports the average
$\ell_2$ estimation error
$E_k=\lVert\theta_k-\theta^\star\rVert_2$
over 100 independent runs. Figure~\ref{fig:sim-1} illustrates how batching and
quantization precision affect estimation accuracy.
In the left panel, larger batches reduce the terminal
error while keeping both the number of transmissions
and the message length fixed, at the cost of additional
samples. In the middle panel, increasing the precision reduces
the gap to unquantized TD; with $m=16$, the quantized
and unquantized curves nearly overlap. Finally, the right panel provides empirical support for the statistical rate predicted by Theorem~\ref{thm:stochastic}, with terminal error $E_K=\lVert\theta_K-\theta^\star\rVert_2$ decaying approximately as $1/\sqrt{T}$ for $m=\{12,16\}$.
\section{Main Convergence Analysis}
\label{sec:convergence}
To prove Theorem~\ref{thm:stochastic}, we first bound the estimation errors from batched estimation of the TD update direction in each epoch. To that end, recall that $\epsilon_H=C_\delta/\sqrt H$, and 
$\eta=\sqrt d/2^m$. 
For a generic transition sample $(s,r,s')$, let
$B:=\phi(s)(\phi(s)-\gamma\phi(s'))^\top$.
Due to feature normalization,
$\norm{B}
\le\norm{\phi(s)}
\bigl(\norm{\phi(s)}+\gamma\norm{\phi(s')}\bigr)
\le2$.
Consequently, since $A=\E[B]$, we have
$\norm{A}\le\E[\norm{B}]\le2$. For each epoch $k \in [K-1]$, define the following:
\begin{equation}\label{eq:ghat}
\widehat g_k(\theta):=\widehat b_k-\widehat A_k\theta,
\qquad \theta\in\R^d,
\end{equation}
where
$\widehat A_k:=H^{-1}\sum_{j\in\mathcal I_k}\phi(s_j)
\bigl(\phi(s_j)-\gamma\phi(s'_j)\bigr)^\top$
and
$\widehat b_k:=H^{-1}\sum_{j\in\mathcal I_k}r_j\phi(s_j)$.
Then $Y_k=\widehat g_k(\theta_k)$ by \eqref{eq:Y}.
 Also, under the i.i.d.\ stationary sampling model, we have 
$\E[\widehat A_k]=A$ and $\E[\widehat b_k]=b$. We have the following result. 

\begin{lemma}\textbf{(Operator Estimation)}
\label{lem:concentration}
Define
\begin{equation}\label{eq:good-event}
\mathcal G:=
\bigcap_{k=0}^{K-1}
\left\{
\norm{\widehat A_k-A}\le\epsilon_H,\,
\norm{\widehat b_k-b}\le\epsilon_H
\right\}.
\end{equation}
Then $\mathbb P(\mathcal G)\ge1-\delta$, and on $\mathcal G$,
\begin{equation}\label{eq:uniform-noise}
\norm{\widehat g_k(\theta)-\gbar(\theta)}
\le\epsilon_H(\norm{\theta}+1)
\end{equation}
simultaneously for all $k=0,\ldots,K-1$ and $\forall \theta\in\R^d$.
\end{lemma}
\begin{proof}
We first control the empirical matrices. For $j\in\mathcal I_k$,
write $B_j:=\phi(s_j)(\phi(s_j)-\gamma\phi(s'_j))^\top$.
Feature normalization gives
\(
\norm{B_j}
\le
\norm{\phi(s_j)}
\bigl(\norm{\phi(s_j)}+\gamma\norm{\phi(s'_j)}\bigr)
\le1+\gamma\le2.
\)
Hence, $\norm{\widehat A_k}\le2$ for every realized batch
and $\E[B_j]=A$.
Moreover, every entry satisfies
$|[B_j]_{i\ell}|
=|\phi_i(s_j)|\,|\phi_\ell(s_j)-\gamma\phi_\ell(s'_j)|
\le2$. Since $\widehat A_k$ averages $H$
independent matrices with expectation $A$, scalar Hoeffding's
inequality~\cite{chungconc} yields, for every $i,\ell\in\{1,\ldots,d\}$,
\[
\mathbb P\!\left(
\bigl|[\widehat A_k-A]_{i\ell}\bigr|
>\frac{\epsilon_H}{d}
\right)
\le
2\exp\!\left(
-\frac{H\epsilon_H^2}{8d^2}
\right)
\le\frac{\delta}{2Kd^2}.
\]
The last inequality follows from
$H\epsilon_H^2=C_\delta^2$ and \eqref{eq:Cdelta}.
A union bound over all $d^2$ entries and $K$ epochs
then ensures that, with probability at least $1-\delta/2$,
\[
\norm{\widehat A_k-A}
\le\norm{\widehat A_k-A}_{\mathrm F}
\le d\max_{i,\ell}
\bigl|[\widehat A_k-A]_{i\ell}\bigr|
\le\epsilon_H
\]
simultaneously for $k=0,\ldots,K-1$. Here, 
$\norm{\cdot}_{\mathrm F}$ denotes the Frobenius norm. Next, consider the empirical reward vectors. Let $(s,r,s')$
be a generic stationary transition tuple and fix
$i\in\{1,\ldots,d\}$. Since
$\E[r|s]=R_\mu(s)$, we have
$[b]_i=\E[R_\mu(s)\phi_i(s)]$. For every $\lambda\in\R$, define
$(\textcolor{blue}{\bullet})\triangleq
\E\!\left[\exp~\!\bigl(\lambda(r\phi_i(s)-[b]_i)\bigr)\right]$.
Conditioning on $s \in \mc{S}$ and using the reward assumption
in Section~\ref{sec:model}, we obtain:
\begin{equation}\medmath{
\begin{aligned}
&(\textcolor{blue}{\bullet})=
\E\!\left[
\exp~\!\bigl(\lambda(R_\mu(s)\phi_i(s)-[b]_i)\bigr)
\E\!\left[
\exp~\!\bigl(\lambda\phi_i(s)(r-R_\mu(s))\bigr)
\,\middle|\,s
\right]
\right]\\
&\overset{\mathrm{(i)}}{\le}
\exp~\!\left(\frac{\lambda^2\sigma^2}{2}\right)
\E\!\left[
\exp~\!\bigl(\lambda(R_\mu(s)\phi_i(s)-[b]_i)\bigr)
\right]\\
&\overset{\mathrm{(ii)}}{\le}
\exp\!\left(
\frac{\lambda^2(\bar r^2+\sigma^2)}{2}
\right).
\end{aligned}}
\end{equation}
Here, (i) follows by applying the conditional sub-Gaussian
assumption to $r-R_\mu(s)$ with scalar $\lambda\phi_i(s)$ and
using $|\phi_i(s)|\le1$. For (ii), Hoeffding's lemma applies to
$R_\mu(s)\phi_i(s)$, which lies in $[-\bar r,\bar r]$ and has mean
$\E[R_\mu(s)\phi_i(s)]=[b]_i$. Therefore,
$r\phi_i(s)-[b]_i$ is sub-Gaussian with variance proxy
$\bar r^2+\sigma^2$. Independence within each batch implies that
$[\widehat b_k-b]_i$ is sub-Gaussian with variance proxy
$(\bar r^2+\sigma^2)/H$.
By \eqref{eq:Cdelta},
$\epsilon_H/\sqrt d
\ge\sqrt{2(\bar r^2+\sigma^2)\log(4Kd/\delta)/H}$.
Scalar sub-Gaussian concentration therefore gives
\[
\mathbb P\!\left(
\bigl|[\widehat b_k-b]_i\bigr|
>\frac{\epsilon_H}{\sqrt d}
\right)
\le
2\exp\!\left(
-\log\!\left(\frac{4Kd}{\delta}\right)
\right)
=\frac{\delta}{2Kd},
\]
again by \eqref{eq:Cdelta}. A union bound over the $d$ coordinates and $K$ epochs,
followed by $\norm{x}\le\sqrt d\max_i|x_i|$, gives
$\norm{\widehat b_k-b}
\le\sqrt d\max_i|[\widehat b_k-b]_i|
\le\epsilon_H$ simultaneously for all $k=0,\ldots,K-1$,
with probability at least $1-\delta/2$.
Combining the matrix and vector bounds yields
$\mathbb P(\mathcal G)\ge1-\delta$.
On $\mathcal G$, for every epoch $k$ and every $\theta\in\R^d$,
the identity
$\widehat g_k(\theta)-\gbar(\theta)
=(\widehat b_k-b)-(\widehat A_k-A)\theta$
and the triangle inequality gives 
$\norm{\widehat g_k(\theta)-\gbar(\theta)}
\le\norm{\widehat b_k-b}+\norm{\widehat A_k-A}\norm{\theta}
\le\epsilon_H(\norm{\theta}+1)$.
\end{proof}

Next, we show that with high probability, every innovation lies within the encoding radius
specified by \eqref{eq:init} and \eqref{eq:range-update}.

\begin{lemma}\textbf{(No Saturation)}
\label{lem:no-saturation}
On the event $\mathcal G$ defined in \eqref{eq:good-event},
\eqref{eq:init} and \eqref{eq:range-update} imply
$\norm{u_k}\le \rho_k$ for all $k=0,\ldots,K-1$. Therefore, by
Lemma~\ref{lem:quant-error}, the reconstruction error satisfies
\begin{equation}\label{eq:q-bound}
q_k=Q_{m,\rho_k}(u_k)-u_k,\qquad
\norm{q_k}\le\eta \rho_k.
\end{equation}
\end{lemma}

\begin{proof}
We prove the claim on the event $\mathcal G$ defined
in \eqref{eq:good-event}. The reward bound
$|R_\mu(s)|\le\bar r$ and the feature bound
$\norm{\phi(s)}\le1$ from Section~\ref{sec:model} give
$\norm{b}
=\norm{\E[R_\mu(s)\phi(s)]}
\le\E[|R_\mu(s)|\norm{\phi(s)}]
\le\bar r$.

Since $h_{-1}=0$ and $Y_0=\widehat g_0(\theta_0)$,
the initial innovation is $u_0=\widehat g_0(\theta_0)$.
Applying \eqref{eq:uniform-noise} on $\mathcal G$ gives
$\norm{u_0}
\le\norm{\gbar(\theta_0)}+\epsilon_H(\norm{\theta_0}+1)
\le\bar r+2\norm{\theta_0}
+\epsilon_H(\norm{\theta_0}+1)
=\rho_0$.
Thus, saturation is inactive at epoch $0$. Suppose $\norm{u_j}\le\rho_j$ for all $j=0,\ldots,k$,
where $k\in\{0,\ldots,K-1\}$.
Since $\norm{u_k}\le\rho_k$,
we have $\sat_{\rho_k}(u_k)=u_k$.
Using $u_k=Y_k-h_{k-1}$, with $Y_k$ defined
in \eqref{eq:Y}, the reconstruction
update~\eqref{eq:h-update} gives
\(
q_k=h_k-Y_k=Q_{m,\rho_k}(u_k)-u_k.
\)
Lemma~\ref{lem:quant-error} then implies
$\norm{q_k}\le\eta\rho_k$.
For $k\le K-2$, we express \(u_{k+1}\) as follows:
\begin{align*}
&u_{k+1}=Y_{k+1}-h_k=Y_{k+1}-Y_k-q_k\\
&
=\widehat g_{k+1}(\theta_{k+1})
  -\widehat g_k(\theta_k)-q_k\\
&=\widehat g_{k+1}(\theta_{k+1})
  -\widehat g_{k+1}(\theta_k) +\widehat g_{k+1}(\theta_k)
  -\widehat g_k(\theta_k)-q_k\\
&=\widehat A_{k+1}(\theta_{k}-\theta_{k+1}) -(\widehat A_{k+1}-\widehat A_k)\theta_k
  +(\widehat b_{k+1}-\widehat b_k)-q_k.
\end{align*}
This decomposition separates the change in parameter from the difference between the two batch estimates evaluated at $\theta_k$. On $\mathcal G$, $\norm{\widehat A_{k+1}-\widehat A_k}\le2\epsilon_H$ and $\norm{\widehat b_{k+1}-\widehat b_k}\le2\epsilon_H$, since each empirical operator is within $\epsilon_H$ of its expectation. Thus, using $\norm{\widehat A_{k+1}}\le2$ and $\theta_{k+1}-\theta_k=\alpha h_k$, we obtain $\norm{u_{k+1}}\le2\norm{\theta_{k+1}-\theta_k}+2\epsilon_H(\norm{\theta_k}+1)+\eta \rho_k=2\alpha\norm{h_k}+2\epsilon_H(\norm{\theta_k}+1)+\eta \rho_k=\rho_{k+1}$. Induction establishes $\norm{u_k}\le \rho_k$ and \eqref{eq:q-bound} for every transmitted innovation.
\end{proof}

Define the sampling error
$\xi_k:=Y_k-\gbar(\theta_k)$. The decoded direction then
admits the decomposition
\(
h_k=\gbar(\theta_k)+\xi_k+q_k.
\)
Here $\xi_k$ accounts for batch estimation and $q_k$ for
finite-bit reconstruction. Recall that $e_k=\theta_k-\theta^\star$.
Since $A\theta^\star=b$, we have
$\gbar(\theta_k)=b-A\theta_k
=-Ae_k$. Our next result provides bounds on the sampling error $\xi_k$, reconstruction error $q_k$, and encoding radius $\rho_k$ as a function of the sub-optimality $e_k$. 
\begin{lemma}\textbf{(Direction and Radius Bounds)}
\label{lem:direction-range}
Suppose $0<\alpha\le\omega/32$. On the event $\mathcal G$ defined in \eqref{eq:good-event}, the following
bounds hold simultaneously for every $k=0,\ldots,K-1$:
\begin{align}
&\norm{\xi_k}
\le\epsilon_H\bigl(\norm{e_k}+1+\norm{\theta^\star}\bigr),
\label{eq:xi-bound}\\
&\norm{\xi_k+q_k}^2
\le2\eta^2\rho_k^2+4\epsilon_H^2\norm{e_k}^2
  +4\epsilon_H^2(1+\norm{\theta^\star})^2,
\label{eq:w-bound}\\
&\rho_{k+1}^2\le12\eta^2\rho_k^2
 +(48\alpha^2+96\epsilon_H^2)\norm{e_k}^2
 +96\epsilon_H^2(1+\norm{\theta^\star})^2.
\label{eq:R-square}
\end{align}
\end{lemma}
\begin{proof}
By \eqref{eq:uniform-noise}, on $\mathcal G$,
\(
\norm{\xi_k}
=\norm{\widehat g_k(\theta_k)-\gbar(\theta_k)}
\le\epsilon_H(\norm{\theta_k}+1).
\)
Using $\theta_k=e_k+\theta^\star$ and the triangle inequality
proves \eqref{eq:xi-bound}. Next, apply $\norm{x+y}^2\le2\norm{x}^2+2\norm{y}^2$
to $\xi_k+q_k$. By \eqref{eq:xi-bound} and
\eqref{eq:q-bound},
\begin{align*}
\norm{\xi_k+q_k}^2
&\le2\norm{\xi_k}^2+2\norm{q_k}^2\\
&\le2\epsilon_H^2
  \bigl(\norm{e_k}+1+\norm{\theta^\star}\bigr)^2
  +2\eta^2\rho_k^2\\
&\le4\epsilon_H^2\norm{e_k}^2
  +4\epsilon_H^2(1+\norm{\theta^\star})^2
  +2\eta^2\rho_k^2,
\end{align*}
where the last inequality uses $(a+b)^2\le2a^2+2b^2$.
This proves \eqref{eq:w-bound}. To bound the next radius, substitute
$h_k=\gbar(\theta_k)+\xi_k+q_k$ into \eqref{eq:range-update}.
Using $\norm{q_k}\le\eta \rho_k$ from \eqref{eq:q-bound}
and $\norm{\xi_k}\le\epsilon_H(\norm{\theta_k}+1)$ from
\eqref{eq:uniform-noise}, we obtain
\begin{align*}
\rho_{k+1}
&\le\eta \rho_k
  +2\alpha
   \bigl(\norm{\gbar(\theta_k)}
         +\norm{\xi_k}+\norm{q_k}\bigr)
  +2\epsilon_H(\norm{\theta_k}+1)\\
&\le\eta(1+2\alpha)\rho_k
  +2\alpha\norm{\gbar(\theta_k)}
  +2(1+\alpha)\epsilon_H(\norm{\theta_k}+1).
\end{align*}
Since $\alpha\le\omega/32\le1/32$, we have
$1+2\alpha\le2$ and $2(1+\alpha)\le4$. Using
$\norm{\theta_k}+1\le\norm{e_k}+1+\norm{\theta^\star}$ gives
\begin{equation}\label{eq:R-linear}
\rho_{k+1}
\le{} 2\eta \rho_k+2\alpha\norm{\gbar(\theta_k)} +4\epsilon_H
 \bigl(\norm{e_k}+1+\norm{\theta^\star}\bigr).
\end{equation}
Squaring this bound and applying
$(a+b+c)^2\le3(a^2+b^2+c^2)$ yields the following:
\begin{align*}
\rho_{k+1}^2
&\le12\eta^2\rho_k^2
  +12\alpha^2\norm{\gbar(\theta_k)}^2 +48\epsilon_H^2
  \bigl(\norm{e_k}+1+\norm{\theta^\star}\bigr)^2.
\end{align*}
Using $\gbar(\theta_k)=-Ae_k$ and the operator-norm bound $\norm{A}\le2$
gives $\norm{\gbar(\theta_k)}
=\norm{Ae_k}\le\norm{A}\norm{e_k}
\le2\norm{e_k}$. Also,
\(
\bigl(\norm{e_k}+1+\norm{\theta^\star}\bigr)^2
\le2\norm{e_k}^2+2(1+\norm{\theta^\star})^2.
\) Substituting these bounds proves \eqref{eq:R-square}.
\end{proof}

The final piece we require is the following result from~\cite{bhandari_finite}. 
\begin{equation}\label{eq:td-coercivity}
\langle \theta^* - \theta, \bar{g}(\theta) \rangle \geq \omega \Vert \theta - \theta^* \Vert^2, \forall \theta \in \mathbb{R}^d,
\end{equation}
where recall that $\omega = (1-\gamma) \tilde{\omega} \in (0,1)$, and $\tilde{\omega}$ is the smallest eigenvalue of $\Sigma= \Phi^{\top} D \Phi$. We now combine these bounds through the joint Lyapunov
function $V_k=\norm{e_k}^2+\alpha\rho_k^2$ to prove
Theorem~\ref{thm:stochastic}.

\(\smallsquare\) \textbf{Proof of Theorem~\ref{thm:stochastic}.}
By Lemma~\ref{lem:concentration}, the event $\mathcal G$
defined in \eqref{eq:good-event} has probability at least
$1-\delta$. We prove the claim on this event.
Recall that $\eta=\sqrt d/2^m$.
The prescribed precision
$m\ge\lceil\frac12\log_2(64d/\omega)\rceil$
gives $\eta^2=d/2^{2m}\le\omega/64$.
Since $T=KH$, the sample-size condition implies
$H\ge c_2C_\delta^2/\omega^2$.
Taking the universal constant $c_2\ge1024$ therefore
ensures that
$\epsilon_H^2=C_\delta^2/H\le\omega^2/1024$.
The stepsize satisfies $0<\alpha\le\omega/32$.
Lemma~\ref{lem:no-saturation} gives
$\norm{u_k}\le\rho_k$ and $\norm{q_k}\le\eta\rho_k$
for every $k=0,\ldots,K-1$.
The sampling and reconstruction errors thus satisfy
the bounds in Lemma~\ref{lem:direction-range}.

We first derive a recursion for
$V_k=\norm{e_k}^2+\alpha\rho_k^2$.
The parameter update and the decomposition of $h_k$ give
\(
e_{k+1}
=e_k+\alpha\gbar(\theta_k)+\alpha(\xi_k+q_k).
\)
From this and~\eqref{eq:td-coercivity}, we obtain
\begin{align*}
\norm{e_{k+1}}^2
\le{}&(1-2\alpha\omega)\norm{e_k}^2
+2\alpha\langle e_k,\xi_k+q_k\rangle\\
&+2\alpha^2\norm{\gbar(\theta_k)}^2
+2\alpha^2\norm{\xi_k+q_k}^2.
\end{align*}
Young's inequality gives
\(
2\langle e_k,\xi_k+q_k\rangle
\le\frac{\omega}{2}\norm{e_k}^2
+\frac{2}{\omega}\norm{\xi_k+q_k}^2.
\)
Together with $\norm{\gbar(\theta_k)}\le2\norm{e_k}$,
this yields
\begin{align*}
\norm{e_{k+1}}^2
\le{}&
\left(1-\frac32\alpha\omega+8\alpha^2\right)
\norm{e_k}^2 +2\left(\frac{\alpha}{\omega}+\alpha^2\right)
\norm{\xi_k+q_k}^2.
\end{align*}
Because $\alpha\omega\le1$, the coefficient of the final
term is at most $4\alpha/\omega$.
Substituting \eqref{eq:w-bound} therefore gives
\begin{equation}\label{eq:e-square}
\begin{split}
\norm{e_{k+1}}^2
\le{}&
\left(1-\frac32\alpha\omega+8\alpha^2
+\frac{16\alpha\epsilon_H^2}{\omega}\right)
\norm{e_k}^2\\
&+\frac{8\alpha\eta^2}{\omega}\rho_k^2
+\frac{16\alpha\epsilon_H^2}{\omega}
(1+\norm{\theta^\star})^2.
\end{split}
\end{equation}
Adding $\alpha$ times \eqref{eq:R-square} to
\eqref{eq:e-square} and collecting terms gives
\begin{equation}\label{eq:V-before}\medmath{
\begin{aligned}
V_{k+1}\le{}&
\left[
1-\frac32\alpha\omega
+\alpha\left(
8\alpha+48\alpha^2+
\left(\frac{16}{\omega}+96\right)\epsilon_H^2
\right)
\right]\norm{e_k}^2\\
&+\alpha\eta^2\left(\frac8\omega+12\right)\rho_k^2
+\alpha\left(\frac{16}{\omega}+96\right)\epsilon_H^2
(1+\norm{\theta^\star})^2.
\end{aligned}}
\end{equation}
Since $0<\omega\le1$ and $\alpha\le\omega/32$,
we have $\alpha\omega\le1/32$ and
\(
\eta^2\left(8/\omega+12\right)
\le
\frac{\omega}{64}\left(8/\omega+12\right)
=
\frac18+\frac{3\omega}{16}
\le\frac5{16}
\le\frac{31}{32}
\le1-\alpha\omega.
\)
Moreover, the stepsize and batch-size bounds give
\(
8\alpha+48\alpha^2+
(16/\omega+96)\epsilon_H^2
\le13\omega/32\le\omega/2.
\)
Thus, the coefficients of $\norm{e_k}^2$ and
$\alpha\rho_k^2$ in \eqref{eq:V-before}
are both at most $1-\alpha\omega$.
Finally, $16/\omega+96\le112/\omega$ bounds the additive
term by
$\mc{O}\bigl(\alpha\epsilon_H^2
(1+\norm{\theta^\star})^2/\omega\bigr)$.
Substituting these estimates into \eqref{eq:V-before}
and using $V_k=\norm{e_k}^2+\alpha\rho_k^2$ yields
\begin{equation}\label{eq:V-recursion}
V_{k+1}
\le(1-\alpha\omega)V_k
+\mc{O}\left(\frac{\alpha\epsilon_H^2}{\omega}\right)
(1+\norm{\theta^\star})^2
\end{equation}
for every $k=0,\ldots,K-1$. Iterating \eqref{eq:V-recursion} gives
\begin{equation}\label{eq:V-unrolled}
\begin{aligned}
V_k
\le{}&(1-\alpha\omega)^kV_0
+\mc{O}\left(\frac{\alpha\epsilon_H^2}{\omega}\right)
(1+\norm{\theta^\star})^2
\sum_{j=0}^{k-1}(1-\alpha\omega)^j
\end{aligned}
\end{equation}
for every $k=1,\ldots,K$.
Since $0<\alpha\omega<1$, the geometric sum is bounded
from above by $1/(\alpha\omega)$.
Using $\epsilon_H^2=C_\delta^2/H$ in
\eqref{eq:V-unrolled} and dropping the nonnegative term
$\alpha\rho_k^2$, we obtain
\begin{equation}\label{eq:parameter-explicit}
\begin{aligned}
\norm{\theta_k-\theta^\star}^2
\le{}&(1-\alpha\omega)^kV_0
+\mc{O}\left(
\frac{C_\delta^2(1+\norm{\theta^\star})^2}{\omega^2H}
\right).
\end{aligned}
\end{equation}
The same bound holds at $k=0$ because
$\norm{\theta_0-\theta^\star}^2\le V_0$.
Thus, \eqref{eq:parameter-explicit} holds simultaneously
for all $k=0,\ldots,K$ on the single event $\mathcal G$;
no additional union bound is needed.

To express the statistical contribution in terms of the
sample budget, substitute $H=T/K$ into
\eqref{eq:parameter-explicit}. This gives
$\mc{O}\!\left(
C_\delta^2(1+\norm{\theta^\star})^2K/(\omega^2T)
\right)$.
Under the prescribed choice
$K=\lceil c_1\log T/(\alpha\omega)\rceil$, this becomes
$\mc{O}\!\left(
C_\delta^2(1+\norm{\theta^\star})^2\log T/
(\alpha\omega^3T)
\right)$, proving \eqref{eq:main-bound}.\\

\section{Conclusion}
We studied policy evaluation with linear function
approximation over a rate-limited channel and proposed
\textmd{\texttt{BAQ-TD}}, which combines model-based variance reduction with
adaptive quantization of innovations.
Our high-probability guarantee matches the unquantized
\td rate, requiring a modest logarithmic-in-sample size communication budget. 
\bibliographystyle{IEEEtran} 
\bibliography{refs}
\end{document}